\documentclass{amsart}

\usepackage[foot]{amsaddr} 

\usepackage{fancyhdr}

 \usepackage{graphicx} 
 \usepackage{amsthm}
 \usepackage[small]{caption}
 \newtheorem{theo}{Theorem}
 \newtheorem*{proof*}{Proof}

\numberwithin{equation}{section}

\begin{document}

\title{Adaptive Learning Rate via Covariance Matrix Based Preconditioning for Deep Neural Networks}


\author{Yasutoshi Ida}
\address{NTT Software Innovation Center, 3-9-11, Midori-cho Musashino-shi, Tokyo, Japan}
\email{ida.yasutoshi@lab.ntt.co.jp}

\author{Yasuhiro Fujiwara}
\email{fujiwara.yasuhiro@lab.ntt.co.jp}
\author{Sotetsu Iwamura}
\email{iwamura.sotetsu@lab.ntt.co.jp}

\keywords{}


\dedicatory{}

\maketitle

\thispagestyle{myheadings}
\markboth{}{}
\pagestyle{plain}
\thispagestyle{fancy}

\begin{abstract}
Adaptive learning rate algorithms such as RMSProp are widely used for training deep neural networks.
RMSProp offers efficient training since it uses first order gradients to approximate Hessian-based preconditioning.
However, since the first order gradients include noise caused by stochastic optimization, the approximation may be inaccurate.
In this paper, we propose a novel adaptive learning rate algorithm called SDProp.
Its key idea is effective handling of the noise by preconditioning based on covariance matrix.
For various neural networks, our approach is more efficient and effective than RMSProp and its variant.
\end{abstract}

\section{Introduction}
Adaptive learning rate algorithms are widely used for the efficient training of deep neural networks.
RMSProp \cite{tieleman2012lecture} and its follow-on methods \cite{zeiler2012adadelta,kingma2014adam} are being used in many deep neural networks such as Convolutional Neural Networks (CNNs) \cite{lecun1998gradient} since they can be easily implemented with high memory efficiency.

The empirical success of RMSProp could be explained by using Hessian-based preconditioning \cite{NIPS2015_5870}.
Hessian is the matrix that represents the curvature of the loss function; Hessian-based preconditioning locally changes the curvature of the loss function.
When training deep neural networks, pathological curvatures such as saddle points \cite{NIPS2014_5486} and cliffs \cite{DBLPPascanuMB13} can slow the progress of first order gradient descent, such as Stochastic Gradient Descent (SGD) \cite{robbins1951stochastic}.
Hessian-based preconditioning improves the condition of the curvature, and thus enhances SGD speed.
However, SGD with Hessian-based preconditioning incurs high computation cost because it generally computes the inverse matrix of Hessian.
Since RMSProp approximates Hessian-based preconditioning by using first order gradients \cite{NIPS2015_5870}, it achieves efficient training.
In addition, RMSProp is easy to implement.
Therefore, in terms of practical use, RMSProp and its variants such as AdaDelta \cite{zeiler2012adadelta} and Adam \cite{kingma2014adam} are still seen as the most powerful approach to training deep neural networks.

However, the first order gradients used in RMSProp include noise caused by stochastic optimization techniques such as mini-batch setting.
With batch setting, since the model inputs are fixed in each iteration, only parameter updates change the gradients.
On the other hand, with mini-batch setting, since the inputs are not fixed in each iteration, gradients can also be changed by randomly selecting the inputs in each iteration.
This change in the mini-batch setting can be seen as noise.
Since RMSProp uses the noisy first order gradients to approximate Hessian-based preconditioning, the approximation may be inaccurate.
This indicates that the efficiency of RMSProp can be improved by effectively handling the noise in the first order gradients.

This paper  proposes a novel adaptive learning rate algorithm called \textit{SDProp}.
The key idea is to use covariance matrix based preconditioning instead of Hessian-based preconditioning.
The covariance matrix is derived by assuming a distribution for the noise in the observed gradients.
Since the distribution effectively captures the noise, SDProp can effectively capture the changes in gradients caused by random input selection in each iteration.
Interestingly, our theoretical analysis reveals that SDProp uses the information of directions over past gradients in adapting the learning rate while RMSProp and its variants use the magnitudes of the gradients.
In experiments, we compare SDProp with RMSProp.
SDProp needs 50 \(\%\)  fewer training iterations than RMSProp to reach the final training loss for CNN in Cifar-10, Cifar-100 and MNIST datasets.
In addition, SDProp outperforms Adam, a state-of-the-art algorithm based on RMSProp, in several datasets.
Our approach is also more effective than RMSProp for training Recurrent Neural Network (RNN) \cite{elman1990finding}  and very deep fully-connected neural networks.

\section{Preliminary}
We briefly review the background of this paper.
First, we describe SGD, which is a basic algorithm in stochastic optimization such as mini-batch setting.
Second, we review RMSProp.
Finally, we explain the relationship between Hessian-based preconditioning and RMSProp.

\subsection{Stochastic Gradient Descent}
\label{sgd}
Many learning algorithms aim at minimizing loss function \(f(\theta)\) with respect to parameter vector, \(\theta\)\cite{l1graph,fastlasso}.
SGD is a popular algorithm in the mini-batch setting.
To minimize \(f(\theta)\), SGD iteratively updates \(\theta\) with a mini-batch of samples as follows:
\vskip -1em
\begin{equation}
\label{sgd_rule}
\theta_{i,t} = \theta_{i,t-1} - \alpha \nabla f\left(\theta_{i, t-1}; x_{t-1}\right)
\end{equation}
where \(\alpha\) is the learning rate, \(\theta_{i, t}\) is the \(i\)-th element of the parameter vector at time \(t\), \(x_{t-1}\) is the sample or mini-batch at time \(t-1\), and \(\nabla f\left(\theta_{i, t-1}; x_{t-1}\right)\) is the first order gradient with respect to the \(i\)-th parameter given by \(x_{t-1}\).
SGD applies Equation (\ref{sgd_rule}) to each sample or mini-batch while Gradient Descent (GD) applies Equation (\ref{sgd_rule}) to all data in the batch setting.
Although \(\nabla f\left(\theta_{i, t-1}; x_{t-1}\right)\) includes noise due to the random selection of mini-batch \(x_{t-1}\), SGD uses it in the training phase.
Since SGD only uses a part of the data for computing \(\nabla f\left(\theta_{i, t-1}; x_{t-1}\right)\), each iteration has reduced computation cost while memory efficiency is high.

\subsection{RMSProp}
RMSProp is a popular algorithm based on SGD for training neural networks.
AdaDelta and Adam are follow-up methods of RMSProp.
RMSProp rapidly reduces loss function \(f(\theta)\) by adapting the learning rate of SGD.
The updating rule of RMSProp is as follows:
\vskip -1em
\begin{eqnarray}
\label{rms_rule1}
&&  v_{i, t} = \beta v_{i, t-1} + (1-\beta) \nabla f\left(\theta_{i, t-1}; x_{t-1}\right)^{2} \\
\label{rms_rule2}
&& \textstyle \theta_{i, t} = \theta_{i, t-1} - \frac{\alpha}{\sqrt{v_{i, t}}+\epsilon} \nabla f\left(\theta_{i, t-1}; x_{t-1}\right)
\end{eqnarray}
where \(v_{i, t}\) is the moving average of uncentered variance over past first order gradients \(\nabla f\left(\theta_{i, t-1}; x_{t-1}\right)\), \(\beta\) is the decay rate for computing \(v_{i, t}\), and \(\epsilon\) is the small value for the stable computation.
Intuitively, RMSProp divides the learning rate, \(\alpha\), by magnitude \(\sqrt{v_{i, t}}\) of the past first order gradients \(\nabla f\left(\theta_{i, t-1}; x_{t-1}\right)\).
Therefore, if the \(i\)-th parameter has large \(\nabla f\left(\theta_{i, t-1}; x_{t-1}\right)\) values in terms of the magnitude in the past, RMSProp yields a small learning rate because \(\sqrt{v_{i, t}}\) in Equation (\ref{rms_rule2}) is large.
Empirically, this idea efficiently reduces the loss function for deep neural networks.
Follow-up methods such as AdaDelta and Adam are based on this idea.
For the convex optimization, regret analysis can be used to explain the efficiency of the methods \cite{kingma2014adam}.
For non-convex optimization such as deep neural networks, the empirical success of RMSProp could be explained by using Hessian-based preconditioning.
We briefly review the relationship between Hessian-based preconditioning and RMSProp by following \cite{NIPS2015_5870} in the next section.

\subsection{Hessian-based Preconditioning}
Some kind of pathological curvature of the loss function slows the progress of SGD \cite{NIPS2014_5486}.
Therefore, it is important to capture the curvature in order to efficiently train deep neural networks.

Hessian-based preconditioning locally changes the function by using Hessian \(H\), which can capture the curvature of the function.
Hessian is the square matrix of the second order gradients of function \(f(\theta)\) represented by \(H\!\!=\!\!\nabla^{2} f(\theta)\).
The condition number of Hessian estimates the extent to which the curvature is pathological.
Condition number is defined as \(\textstyle \sigma_{{\rm max}}(H) / \sigma_{{\rm min}}(H)\) where \(\sigma_{{\rm max}}(H)\) and \(\sigma_{{\rm min}}(H)\) are the largest and smallest singular values of \(H\), respectively.
The function has less pathological curvature if the condition number has a small value.
This is because the function equally curves if it has small condition number.
Therefore, we can increase the efficiency of the training by reducing the Hessian condition number \cite{NIPS2015_5870}.

Hessian-based preconditioning locally transforms an original parameter into another parameter so that the Hessian has small condition number.
Preconditioning matrix \(D\) gives transformations such as \(\textstyle \hat{\theta}\!\!=\!\!D^{1/2}\theta\) where \(\textstyle \hat{\theta}\) is the transformed parameter.
By using \(\textstyle \hat{\theta}\), function \(f\) is transformed into function \(\textstyle \hat{f}\) where \(\textstyle f(\theta)\!\!= \!\!f(D^{-1/2}\hat{\theta})\!\!=\!\!\hat{f}(\hat{\theta})\).
If \(\textstyle \hat{f}(\hat{\theta})\) has smaller condition number than \(\textstyle f(\theta)\), we can efficiently train a model by applying first order gradient descent to \(\textstyle \hat{\theta}\).
The updating rule of \(\textstyle \hat{\theta}\) is \(\textstyle \hat{\theta}_{t}\!\!=\!\!\hat{\theta}_{t-1}\!\!-\!\!\alpha \nabla \hat{f}(\hat{\theta})\).
Since \(\textstyle \nabla \hat{f}(\hat{\theta})\!\!=\!\!D^{-1/2} \nabla f(\theta)\), we have the following form for original parameter \(\textstyle \theta\):
\vskip -1em
\begin{equation}
\label{updating_D}
\textstyle \theta_{t} = \theta_{t-1} - \alpha D^{-1} \nabla f(\theta_{t-1}).
\end{equation}
If \(\textstyle \hat{H}\) is the Hessian of transformed function \(\textstyle \hat{f}(\hat{\theta})\), \(\textstyle \hat{H}\) is given as \(\textstyle \hat{H}\!\!=\!\!(D^{-1/2})^{\rm T}HD^{-1/2}\).
When \(\textstyle D^{1/2}\!\!=\!\!H^{1/2}\), \(\textstyle \hat{H}\) has a smaller condition number because \(\textstyle \hat{H}\) is an identity matrix.
In this case, Equation (\ref{updating_D}) corresponds the Newton method.
However, \(\textstyle H^{1/2}\) exists only when \(H\) is positive-semidefinite.
Since deep neural networks have many saddle points where Hessian can be indefinite \cite{NIPS2014_5486}, the Newton method is unsuitable for training deep neural networks.
On the other hand, the diagonal equilibration matrix of \(\textstyle D\!\!=\!\!\sqrt{{\rm diag}\left(H^{2}\right)}\) works well even if \(H\) is indefinite \cite{NIPS2015_5870}.
This indicates that GD can efficiently escape from saddle points by preconditioning based on the diagonal equilibration matrix.

In RMSProp, the role of \(\textstyle \sqrt{v_{i,t}}\) in Equation (\ref{rms_rule2}) could be explained by using Hessian-based preconditioning \cite{NIPS2015_5870}.
A comparison of Equation (\ref{updating_D}) to Equation (\ref{rms_rule2}) indicates that \(\textstyle \sqrt{v_{i,t}}\) corresponds to the \(i\)-th element of the diagonal preconditioning matrix.
In addition, empirical results suggest that \(\textstyle \sqrt{v_{i,t}}\) approximates the \(i\)-th element of the diagonal equilibration matrix which can be used to efficiently train deep neural networks \cite{NIPS2015_5870}.
Thus, RMSProp can be interpreted as Hessian-based preconditioning using an approximated diagonal equilibration matrix in the mini-bath setting.
Therefore, since RMSProp is more efficient in escaping from saddle points than SGD, RMSProp and its follow-up methods achieve high efficiency.

\section{Proposed Method}
We first introduce the novel preconditioning idea.
Then, we derive SDProp based on this idea.

\subsection{Idea}
RMSProp approximates Hessian-based preconditioning by using the first order gradients \(\nabla f\left(\theta_{i, t-1}; x_{t-1}\right)\) as described in the preliminary section.
However, in stochastic optimization approaches such as mini-batch setting, the first order gradients \(\nabla f\left(\theta_{i, t-1}; x_{t-1}\right)\) include noise because input \(x_{t-1}\) is randomly selected in each iteration.
Since the first order gradients \(\nabla f\left(\theta_{i, t-1}; x_{t-1}\right)\) in Equation (\ref{rms_rule1}) and the square roots of the uncentered variances \(\sqrt{v_{i,t}}\) in Equation (\ref{rms_rule2}) contain noise, it is difficult to effectively approximate Hessian-based preconditioning.
In order to effectively handle the noise, we replace Hessian-based preconditioning with covariance matrix based preconditioning.

In covariance matrix based preconditioning, we assume that the first order gradients \(\nabla f\left(\theta_{i, t-1}; x_{t-1}\right)\) follow a Gaussian distribution.
This is because the field of probabilistic modeling uses Gaussian distributions to model the noise of observations\cite{sra2012optimization,dditsm,bayesiannn,tprocess}.
By following \cite{sra2012optimization}, we assume the following Gaussian distribution of first order gradient \(\hat{g}_{t} = \nabla f\left(\theta_{t-1}; x_{t-1}\right) \in \mathcal{R}^{d}\):
\vskip -1em
\begin{eqnarray}
\label{likelihood}
\hat{g}_{t}|\bar{g}_{t} \sim N(\bar{g}_{t},C_{t})
\end{eqnarray}
where \(\bar{g}_{t} \in \mathcal{R}^{d}\) is the true gradient without the noise while \(\hat{g}_{t} \in \mathcal{R}^{d}\) includes the noise.
\(N(\bar{g}_{t},C_{t})\) is a Gaussian distribution with mean \(\bar{g}_{t}\) and covariance matrix \(C_{t}\); \(C_{t}\) is the covariance matrix of \(\hat{g}_{t}\) whose size is \(d \times d\).
The diagonal elements in \(C_{t}\) represent the magnitude of oscillation of the first order gradients \(\hat{g}_{t}\) that include the noise.
Specifically, let \(C_{t}[i,j]\) be the \(i\)-th row and the \(j\)-th column element in \(C_{t}\), \(C_{t}[i,j]\) represents the covariance of the \(i\)-th and the \(j\)-th first order gradient.
Therefore, if the \(i\)-th first order gradient strongly correlates with the \(j\)-th first order gradient, \(C_{t}[i,j]\) has large absolute value.
On the other hand, \(C_{t}[i,i]\) represents the variance of the \(i\)-th first order gradient.
Therefore, \(C_{t}[i,i]\) has large value if the first order gradient strongly oscillates in the \(i\)-th dimension.

Intuitively, large oscillations in $i$-th dimension incur high variance of updating directions and inefficient progress in plain SGD.
However, it is difficult to reduce the oscillation since it can be a result of the noise induced by the mini-batch setting.
How can we reduce the oscillation by using \(C_{t}\) ?
This is the motivation behind our approach; plain SGD efficiently progresses if we can control the oscillation by utilizing \(C_{t}\).
In this paper, we propose the preconditioning of \(C_{t}\) to control the oscillation.
While Hessian-based preconditioning reduces the condition number of Hessian, our preconditioning reduces the condition number of \(C_{t}\) by transforming \(C_{t}\) into an identity matrix.
We describe our approach in the next section.

\subsection{Covariance Matrix Based Preconditioning}
The previous section suggests that large values in the diagonal of \(C_{t}\) prevent the efficient progress of SGD.
Therefore, if we could control the values in the diagonal of \(C_{t}\), we improve the efficiency of SGD.
Our covariance matrix based preconditioning transforms \(C_{t}\) into \(\rho^{2}\textit{ I}\) where \(\textit{ I}\) is an identity matrix whose size is \(d \times d\) and \(\rho\) is a hyper-parameter that has a positive value.
Since the element in the diagonal of \(C_{t}\) represents the variance of first order gradient, we can hold the variance to constant value \(\rho^{2}\).
If the variance is larger than \(\rho^{2}\), its value is reduced to \(\rho^{2}\).
Therefore, SGD efficiently progresses if we transform \(C_{t}\) into \(\rho^{2}\textit{ I}\).

We first describe the approach used to transform \(C_{t}\) into \(\textit{ I}\) instead of \(\rho^{2}\textit{ I}\).
This is because once \(C_{t}\) is transformed into \(\textit{ I}\), it is easy to transform \(\textit{ I}\) into \(\rho^{2}\textit{ I}\) as we describe later.
Hessian-based preconditioning transforms first order gradients to yield \(\textstyle \nabla \hat{f}(\hat{\theta})\!\!=\!\!D^{-1/2} \nabla f(\theta)\) where \(D\) is a preconditioning matrix.
The preconditioning matrix of \(D^{1/2}=H^{1/2}\) reduces the condition number of Hessian \(H\) as described in the preliminary section.
Unlike the previous approach, we execute the preconditioning of \(C_{t}\) and so use the transformation \(\textstyle g_{p}\!\!=\!\!D^{-1}\hat{g}_{t}\). 
In this transformation, \(g_{p}\) is a transformed first order gradient and \(\hat{g}_{t}\) is a first order gradient as defined in Equation (\ref{likelihood}).
Since the transformation is an affine transformation of \(\hat{g}_{t}\) generated from the Gaussian distribution in Equation (\ref{likelihood}), we have following distribution of \(g_{p}\):
\vskip -1em
\begin{eqnarray}
\label{g_p}
g_{p}=D^{-1}\hat{g}_{t}|\bar{g}_{t} \sim N(D^{-1}\bar{g}_{t},D^{-1}C_{t}(D^{-1})^{{\rm T}}).
\end{eqnarray}
In Equation (\ref{g_p}), we use the following major rule to transform Equation (\ref{likelihood}) into (\ref{g_p}): if \(X\!\!\sim\!\!N(m, \Sigma)\) and \(Y\!\!=\!\!AX\), then \(Y\!\!\sim\!\!N(Am, A \Sigma A^{T})\); \(N(m, \Sigma)\) is a Gaussian distribution that has mean \(m\) and covariance matrix \(\Sigma\), \(A\) is a matrix for affine transformation and \(Y\) is a transformed variable.
By setting \(\textstyle D\!\!=\!\!C_{t}^{1/2}\) in Equation (\ref{g_p}), we have the following property :
\begin{theo}
\label{theorem1}
If we transform first order gradient \(\hat{g}_{t}\) to yield \(\textstyle g_{p}\!\!=\!\!C_{t}^{-1/2}\hat{g}_{t}\), we have the following Gaussian distribution:
\vskip -1em
\begin{eqnarray}
\label{trans_var}
g_{p}|\bar{g}_{t} \sim N\left(C_{t}^{-\frac{1}{2}}\bar{g}_{t}, \textit{ I}\right)
\end{eqnarray}
where \(\textit{ I}\) is an identity matrix whose size is \(d \times d\).
\end{theo}
\begin{proof*}
By using eigen decomposition, we can represent \(C_{t}\) as \(C_{t}=U\Sigma U^{\rm{T}}\) where \(U\) is an orthogonal matrix of \(d \times d\) and \(\Sigma\) is a diagonal matrix of \((\lambda_{1}, \lambda_{2},..., \lambda_{d})\).
Since \(C_{t}\) is assumed to be a positive semi-definite matrix, all eigen values are equal to or higher than 0.
Thus, \(C^{1/2}_{t}\) can be computed as \(C^{1/2}_{t}\!\!=\!\!U\Sigma^{1/2} U^{\rm{T}}\).
By setting the covariance term of Equation (\ref{g_p}) to \(\textstyle D^{-1}\!\!=\!\!(C_{t}^{1/2})^{-1}\!\!=\!\!U\Sigma^{-1/2} U^{\rm{T}}\), the Gaussian distribution of \(g_{p}\) is represented as follows:
\vskip -1em
\begin{eqnarray*}
&&\!\!\!\!\!\!\!\!\!\!\!\!g_{p}=C_{t}^{-1/2}\hat{g}_{t}|\bar{g}_{t} \sim N(C_{t}^{-1/2}\bar{g}_{t}, C_{t}^{-1/2}C_{t}(C_{t}^{-1/2})^{{\rm T}}) \\
&&\!\!\!\!\!\!\!\!\!\!\!\!=N(C_{t}^{-1/2}\bar{g}_{t}, U\Sigma^{-1/2} U^{\rm{T}}U\Sigma U^{\rm{T}}(U\Sigma^{-1/2} U^{\rm{T}})^{{\rm T}}) \\
&&\!\!\!\!\!\!\!\!\!\!\!\!=N(C_{t}^{-1/2}\bar{g}_{t}, \textit{ I}). \\
\end{eqnarray*}
\vskip -1em
In the above formulations, since \(U\) is an orthogonal matrix, we use \(UU^{\rm T}\!\!=\!\!\textit{ I}\) and \((U\Sigma^{-1/2} U^{\rm{T}})^{\rm{T}}\!\!=\!\!U\Sigma^{-1/2} U^{\rm{T}}\).
As a result, we have the distribution of Equation (\ref{trans_var}).
\qed
\end{proof*}
The above theorem indicates that the transformation of \(\textstyle g_{p}=C_{t}^{-1/2}\hat{g}\) results in the Gaussian distribution of \(g_{p}\) whose covariance matrix is identity matrix \(\textit{ I}\).
In other words, we can control the covariance matrix to be \(\textit{I}\) by using \(g_{p}\) instead of \(\hat{g}\).

Our preconditioning transforms the value of variance for first order gradients into 1 by using \(g_{p}\).
However, \(g_{p}\) may have an extremely large value if the variance is 1.
Thus, we introduce hyper-parameter \(\rho\) to generalize our preconditioning.
Specifically, by using the transformation of \(\textstyle g_{p}\!\!=\!\!\rho C_{t}^{-1/2}\hat{g}\) instead of \(\textstyle g_{p}\!\!=\!\!C_{t}^{-1/2}\hat{g}\), we have the following distribution:
\begin{eqnarray}
\label{trans_var2}
\rho C_{t}^{-\frac{1}{2}}\hat{g}_{t}|\bar{g}_{t} \sim N\left(\rho C_{t}^{-\frac{1}{2}}\bar{g}_{t}, \rho^{2}\textit{ I}\right).
\end{eqnarray}
The above equation denotes that \(\rho\) controls the value of the covariance matrix while the previous transformation only gives an identity matrix as shown in Equation (\ref{trans_var}).
We show that \(\rho\) has the same role as learning rate \(\alpha\) when we derive SDProp in the next section.

Since we compute the first order gradients at each time \(t\) in SGD, we have to incrementally compute the covariance matrix \(C_{t}\) although Theorem \ref{theorem1} is based on the property that \(C_{t}\) is a positive semi-definite matrix.
In order to incrementally compute \(C_{t}\) as a positive semi-definite matrix, we use the online updating rule of \cite{sra2012optimization} as follows:
\begin{eqnarray}
\label{expcov4}
&C_{t}&\!\!\!\!\!\! =\! \gamma C_{t-1}\!+\! \gamma (1 \!-\! \gamma)(\hat{g}_{t} \!-\! \mu_{t-1})(\hat{g}_{t} \!-\! \mu_{t-1})^{{\rm T}} \\
\label{expcov3}
&\mu_{t}&\!\!\!\!\!\! =\! \gamma \mu_{t-1}\! +\! (1\! -\! \gamma ) \hat{g}_{t} 
\end{eqnarray}
where \(\mu_{t}\) is the moving average of \(\hat{g}_{t}\) and \(\gamma\) is the hyper-parameter of the decay rate for the moving average that has \(\gamma \in [0,1)\).
\(C_{t}\) and \(\mu_{t}\) are initialized as \(\mu_{1} = \hat{g}_{1}\) and \(C_{1} = 0\).
The above updating rule gives the following property:
\begin{theo}
\label{theorem2}
If we compute covariance matrix \(C_{t}\) by using Equations (\ref{expcov4}) and (\ref{expcov3}), \(C_{t}\) is positive semi-definite.
\end{theo}
\begin{proof*}
In order to prove Theorem \ref{theorem2}, we first prove that \((\hat{g}_{t} \!-\! \mu_{t-1})(\hat{g}_{t} \!-\! \mu_{t-1})^{{\rm T}}\) in Equation (\ref{expcov4}) is a positive semi-definite matrix.
By setting \(y=x^{\rm T}(\hat{g}_{t} \!-\! \mu_{t-1})\), we have:
\begin{eqnarray*}
&&x^{\rm T}(\hat{g}_{t} \!-\! \mu_{t-1})(\hat{g}_{t} \!-\! \mu_{t-1})^{{\rm T}}x=yy^{\rm T} \geq 0.
\end{eqnarray*}
By following the definition of positive semi-definite matrixes, if we have matrix \(A\) of \(d \times d\) such that \(x^{{\rm T}}Ax \geq 0\) holds for every non-zero column vector \(x\) of \(d\) real numbers, \(A\) is a positive semi-definite matrix.
Since the above inequation shows that \(x^{\rm T}(\hat{g}_{t} \!-\! \mu_{t-1})(\hat{g}_{t} \!-\! \mu_{t-1})^{{\rm T}}x \geq 0\) holds, it is clear that \((\hat{g}_{t} \!-\! \mu_{t-1})(\hat{g}_{t} \!-\! \mu_{t-1})^{{\rm T}}\) is a positive semi-definite matrix even if \(\mu_{t-1}\) in Equation (\ref{expcov3}) has any real value.

Then, we prove that \(C_{t}\) in Equation (\ref{expcov4}) is a positive semi-definite matrix by mathematical induction. \\
Initial step: If \(t\!\!=\!\!1\), the initialization yields \(C_{1}\!=\!0\).
Since \(C_{2}\) is computed as \(C_{2}\!=\!\gamma (1-\gamma)(\hat{g}_{2}-\mu_{1})(\hat{g}_{2}-\mu_{1})^{{\rm T}}\) by using Equation (\ref{expcov4}) and (\ref{expcov3}), \(C_{2}\) is a positive semi-definite matrix.
This is because \((\hat{g}_{t} \!-\! \mu_{t-1})(\hat{g}_{t} \!-\! \mu_{t-1})^{{\rm T}}\) is a positive semi-definite matrix as proved above.\\
Inductive step: We assume that \(C_{t-1}\) is a positive semi-definite matrix.
Since \(C_{t}\) is computed as \(C_{t}\!=\!\gamma C_{t-1}+\gamma (1-\gamma)(\hat{g}_{t}-\mu_{t-1})(\hat{g}_{t}-\mu_{t-1})^{{\rm T}}\) by using Equations (\ref{expcov4}) and (\ref{expcov3}), \(x^{{\rm T}}C_{t}x\) is represented as follows:
\begin{eqnarray*}
&&\!\!\!\!\!\!\!\!\!\!\!\!\!x^{{\rm T}}C_{t}x=x^{{\rm T}}(\gamma C_{t-1}\!+\! \gamma (1 \!-\! \gamma)(\hat{g}_{t} \!-\! \mu_{t-1})(\hat{g}_{t} \!-\! \mu_{t-1})^{{\rm T}})x \\ \nonumber
&&=\gamma x^{{\rm T}}C_{t-1}x\!+\! \gamma (1 \!-\! \gamma)x^{{\rm T}}(\hat{g}_{t} \!-\! \mu_{t-1})(\hat{g}_{t} \!-\! \mu_{t-1})^{{\rm T}}x.
\end{eqnarray*}
In the above equation, \(x^{{\rm T}}C_{t-1}x \geq 0\) and \(x^{{\rm T}}(\hat{g}_{t} \!-\! \mu_{t-1})(\hat{g}_{t} \!-\! \mu_{t-1})^{{\rm T}}x \geq 0\) because \(C_{t-1}\) and \((\hat{g}_{t} \!-\! \mu_{t-1})(\hat{g}_{t} \!-\! \mu_{t-1})^{{\rm T}}\) are positive semi-definite matrices.
Therefore, \(C_{t}\) is a positive semi-definite matrix because \(x^{{\rm T}}C_{t}x \geq 0\) holds in the above equation.
This completes the inductive step.
\qed
\end{proof*}
Thus, if we compute \(C_{t}\) by using Equations (\ref{expcov4}) and (\ref{expcov3}), we can execute the preconditioning specified by Theorem \ref{theorem1}.

Note that Hessian-based preconditioning cannot control the oscillation of first order gradients.
This is because its transformation results in the distribution of \(N(H^{-\frac{1}{2}}\bar{g}_{t},H^{-\frac{1}{2}}C_{t}(H^{-\frac{1}{2}})^{{\rm T}})\) where the covariance matrix is uncontrollable.
In addition, since Hessian \(H\) may not be a positive semi-definite matrix, it is difficult to compute \(H^{-1/2}\).
Therefore, our covariance matrix based preconditioning inherently differs from Hessian-based preconditioning.
Our idea of preconditioning \(C_{t}\) is more suitable than Hessian-based preconditioning in handling the oscillation triggered by the noise of first order gradients.

\subsection{Algorithm}
Since deep neural networks have a large number of parameters, the idea described in the previous section incurs large memory consumption of \(O(d^{2})\) where \(d\) is the number of parameters.
In addition, it costs \(O(d^{3})\) time to compute \(\textstyle D\!\!=\!\!C_{t}^{1/2}\) by using eigenvalue decomposition \cite{halko2011finding}.
To avoid these problems, we employ diagonal preconditioning matrix \(\textstyle D\!\!=\!\!{\rm diag}(C_{t})^{1/2}\).
Since this approach only needs the diagonal terms, the memory and computation costs are \(O(d)\).
Although this approach ignores the correlation of first order gradients, it is sufficient to control the oscillation in each dimension.
This is because the diagonal of \(C_{t}\) represents the variance of the oscillation as described in the previous section.
By picking the diagonal of Equation (\ref{trans_var2}), the updating rule is:
\vskip -1em
\begin{equation}
\label{initial_updaitng}
\theta_{t} = \theta_{t-1} - \rho \cdot {\rm diag}\left(C_{t}\right)^{-\frac{1}{2}}\nabla f\left(\theta_{i, t-1}; x_{t-1}\right).
\end{equation}
We rewrite this updating rule (all steps) as follows:
\vskip -1em
\begin{eqnarray}
\label{diag_expcov3}
\!\!\!\!\!\!\!\!\mu_{i, t}&\!\!\!\!\!=\!\!\!\!\!&\gamma \mu_{i, t-1}\!+\!(1\!-\! \gamma )  \nabla f\left(\theta_{i, t-1}; x_{t-1}\right) \\
\label{diag_expcov4}
\!\!\!\!\!\!\!\!c^{2}_{i, t}&\!\!\!\!\!=\!\!\!\!\!&\gamma c^{2}_{i, t-1}\!+\! \gamma (1 \!-\! \gamma)( \nabla f\left(\theta_{i, t-1}; x_{t-1}\right) \!-\! \mu_{i, t-1})^{2} \\
\label{diag_expcov5}
\textstyle \!\!\!\!\!\!\!\!\theta_{i,t}&\!\!\!\!\!=\!\!\!\!\!&\theta_{i,t-1}\!-\!\frac{\scriptstyle \rho}{ \scriptstyle \sqrt{c^{2}_{i,t}}\!+\!\epsilon}\nabla f(\theta_{i,t-1};x_{t-1})
\end{eqnarray}
where \(\mu_{i, t}\) is the moving average of first order gradients for the \(i\)-th parameter at time \(t\) and \(\gamma\) is the hyper-parameter of the decay rate for the moving average that has \(\gamma \in [0,1)\).
\(c^{2}_{i, t}\) is the exponentially moving variance of first order gradients for the \(i\)-th parameter at time \(t\).
We use \(\gamma\) in Equation (\ref{diag_expcov4}) as the decay rate of the exponentially moving variance.
\(\mu_{i, t}\) and \(c^{2}_{i, t}\) are initialized as \(\mu_{i, 1}\!\!=\!\!\nabla f\left(\theta_{i, 0}; x_{0}\right)\) and \(c^{2}_{i, 1}\!\!=\!\!0\), respectively.
For stable computation, \(\epsilon\) is set at a small positive value.
Equation (\ref{diag_expcov5}) corresponds to Equation (\ref{initial_updaitng}).
We call the algorithm \textit{ SDProp} because Equation (\ref{diag_expcov5}) includes \textit{ Standard Deviation} \(\scriptstyle \sqrt{c^{2}_{i, t}}\).
Although \(c_{i,t}\) includes the bias imposed by initialization, we can remove the bias in the same way as \cite{kingma2014adam}.

Notice that \(\rho\) takes the same role as learning rate \(\alpha\) in Equation (\ref{rms_rule2}) of RMSProp.
Therefore, Equation (\ref{diag_expcov5}) divides the learning rate by the square root of \textit{centered} variance \(c_{i,t}^{2}\) while Equation (\ref{rms_rule2}) of RMSProp divides the learning rate by the square root of \textit{uncentered} variance \(v_{i,t}^{2}\).
In other words, RMSProp and its follow-up methods such as Adam adapt the learning rate by the magnitude of gradients while we adapt it by the variance of gradients.
Although RMSProp and SDProp have similar updating rules, they have totally different goals as described in the previous sections.
RMSProp executes Hessian-based preconditioning while SDProp executes covariance matrix based preconditioning.

\section{Experiments}
\begin{figure*}[t]
\begin{center}
 \begin{minipage}{0.22\hsize}
  \begin{center}
\includegraphics[viewport=0 0 504 504, width=\columnwidth]{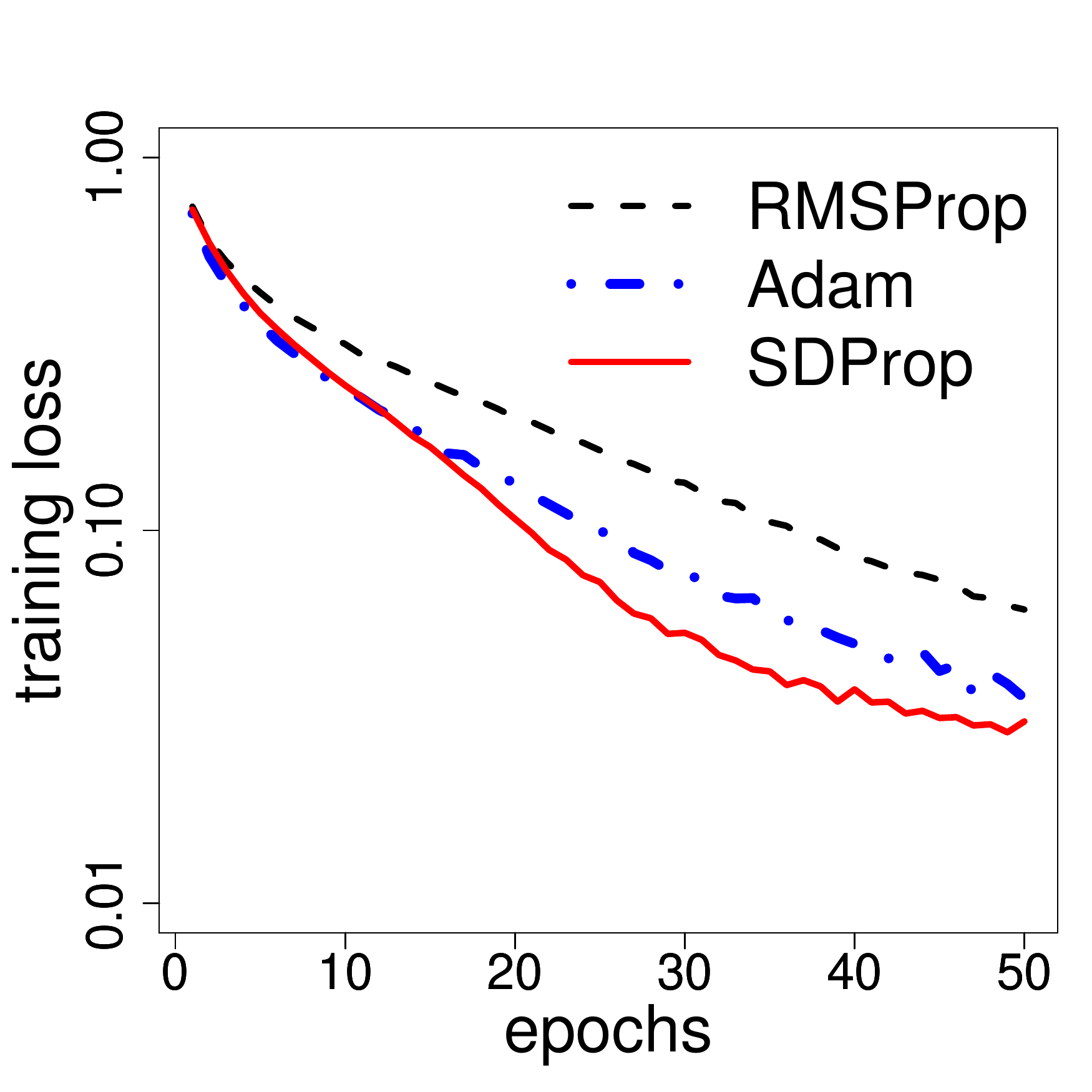} \\
{\small (a) Cifar-10}
  \end{center}
 \end{minipage}
 \begin{minipage}{0.22\hsize}
  \begin{center}
\includegraphics[viewport=0 0 504 504 ,width=\columnwidth]{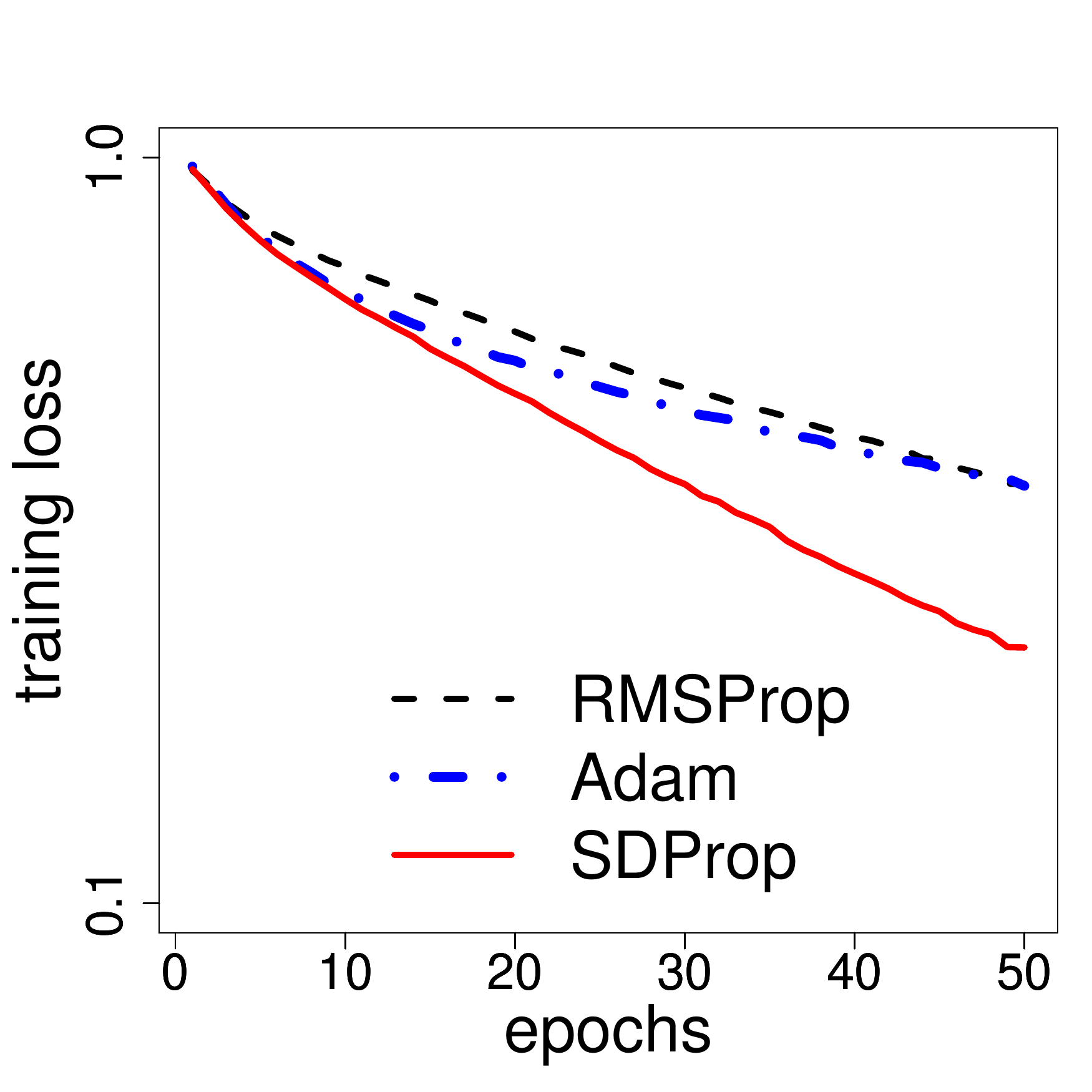} \\
{\small (b) Cifar-100}
  \end{center}
 \end{minipage}
  \begin{minipage}{0.22\hsize}
  \begin{center}
\includegraphics[viewport=0 0 504 504 ,width=\columnwidth]{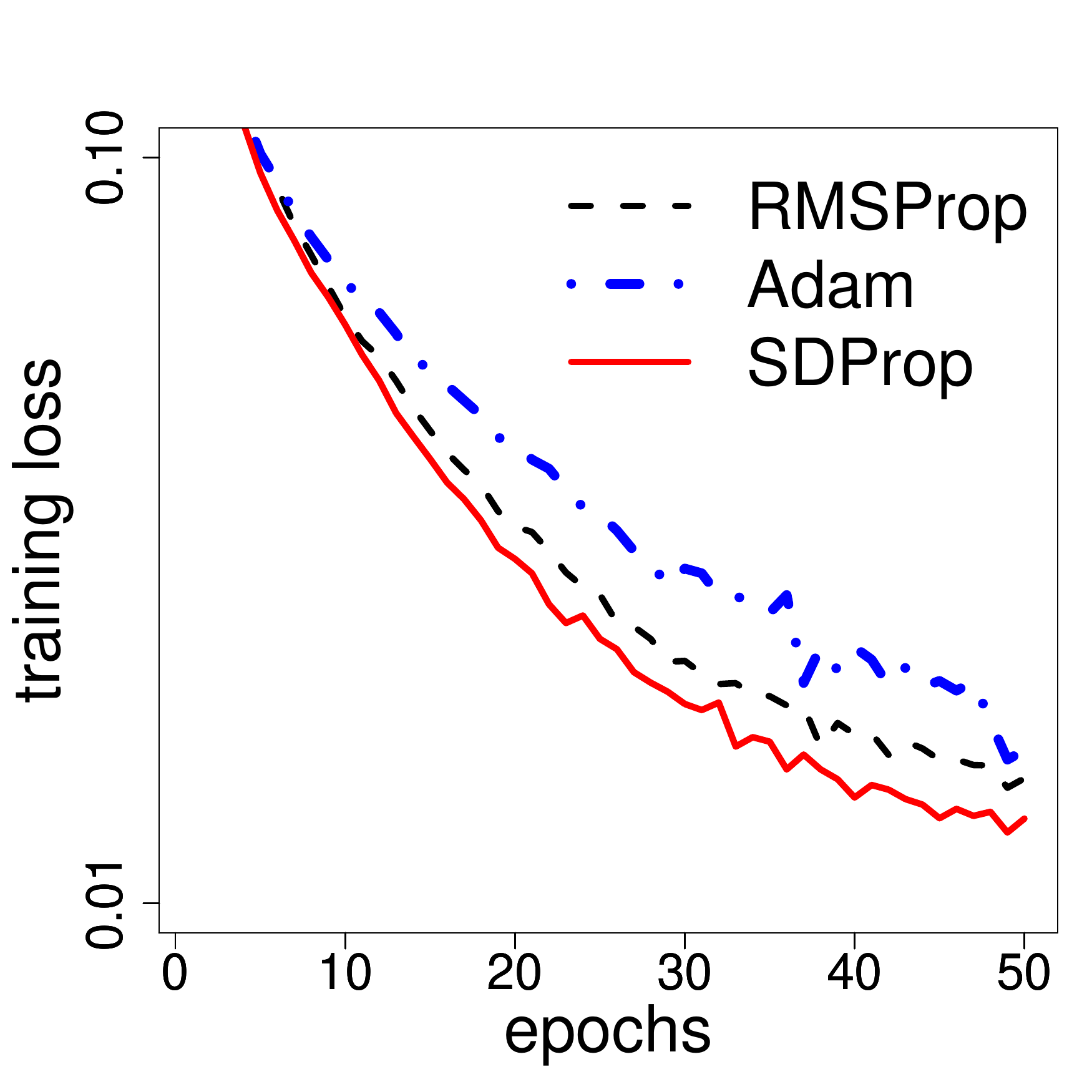} \\
{\small (c) SVHN} 
  \end{center}
 \end{minipage}
  \begin{minipage}{0.22\hsize}
  \begin{center}
\includegraphics[viewport=0 0 504 504 ,width=\columnwidth]{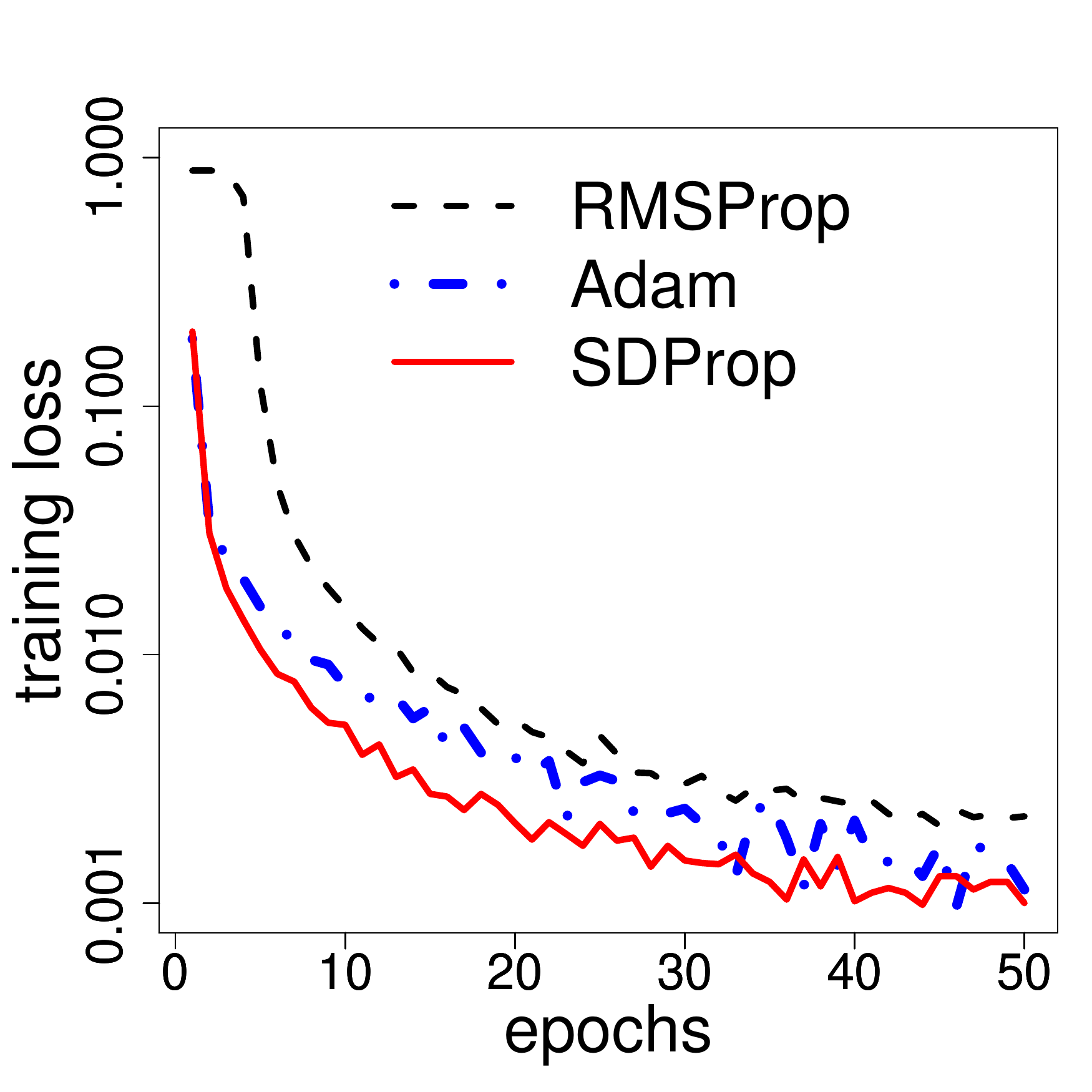} \\
{\small (d) MNIST}
  \end{center}
 \end{minipage}
 \caption{\small Training losses for CNN. We show the results for (a) Cifar-10, (b) Cifar-100, (c) SVHN and (d) MNIST.}
 \label{some_datasets}
\end{center}
\vskip -0.2in
\end{figure*}

We performed experiments to compare SDProp to RMSProp and Adam, a state-of-the-art algorithm based on RMSProp.
 \cite{kingma2014adam} shows that Adam is a more efficient and effective approach than RMSProp or AdaDelta by integrating momentum into RMSProp.
First, we show the efficiency and effectiveness of our approach by using CNN.
Second, since SDProp effectively handles the oscillation described in the previous section, we evaluate SDProp by using small mini-batches which suffer noise in the first order gradients.
Third, we show the efficiency and effectiveness of SDProp for RNN.
Fourth, we demonstrate the effectiveness of SDProp for 20 layered fully-connected neural network that is difficult to train due to many sadle points.

\subsection{Efficiency and Effectiveness for CNN}
We investigate the efficiency and effectiveness of SDProp.
We used 4 datasets to assess the classification of images; Cifar-10, Cifar-100 \cite{krizhevsky2009learning}, SVHN \cite{sermanet2012convolutional} and MNIST.
The experiments were conducted on a 7-layered CNN 
with ReLU activation function.
The loss function was negative log likelihood.
We compared SDProp to RMSProp and Adam.
In SDProp, we tried various combinations of hyper-parameters by using \(\gamma \!\in \!\{0.9, 0.99\}\) and \(\rho \!\in \!\{0.1, 0.01, 0.001\}\).
In RMSProp, we tried combinations of hyper-parameters by using \(\beta \!\in \!\{0.9, 0.99\}\) and \(\alpha \!\in \!\{0.1, 0.01, 0.001\}\).
As a result, SDProp achieves the lowest loss in the settings of \(\gamma=0.99\), \(\rho=0.001\).
RMSProp has the lowest loss when \(\beta=0.99\) and \(\alpha=0.001\).
Adam achieves the lowest loss when \(\beta_{1}=0.9\), \(\beta_{1}=0.999\) and \(\alpha=0.001\).
The mini-batch size was 128.
The number of epochs was 50.
We use the training loss to evaluate the algorithms because they optimize the training criterion.

Figure \ref{some_datasets} shows the training losses of each dataset.
In Cifar-10, Cifar-100 and SVHN, SDProp yielded lower losses than RMSProp and Adam in early epochs.
In MNIST, although the training loss of SDProp and Adam nearly reached 0.0, SDProp reduces the loss faster than Adam.
SDProp needs 50 \(\%\)  fewer training iterations than RMSProp to reach its final training loss in Cifar-10, Cifar-100 and MNIST.
This suggests that our idea of covariance matrix based preconditioning is more efficient and effective than Hessian-based preconditioning in the mini-batch setting because RMSProp and Adam approximate Hessian-based preconditioning as described in the preliminary section.
Since SDProp captures the noise, it effectively reduces the loss even if the gradients are noisy.
In the next experiment, we investigate the performance of SDProp in terms of its effectiveness against noise by using noisy first order gradients.

\subsection{Sensitivity of Mini-batch Size}
\begin{table}[!t]
  \caption{\small Training accuracy percentage for Cifar-10 in CNN for different mini-batch sizes. We tuned the hyper-parameters; the 1st row presents mini-batch size.}
  \vskip 0.1in
  \label{batchsize}
  \centering
  \begin{tabular}{|c|c|c|c|c|c|} \hline
       & \textit{ 16} & \textit{ 32} & \textit{ 64} & \textit{ 128} \\ \hline 
    RMSProp & 81.42 & 93.10 & 94.98 & 95.07 \\  \hline
    Adam & 83.24 & 93.57 & 95.48 & 97.12 \\  \hline
    SDProp & \textbf{ 90.17} & \textbf{ 94.87} & \textbf{ 96.54} & \textbf{ 97.31} \\ \hline
  \end{tabular}
\end{table}

The previous experimental results show that SDProp is more efficient and effective than existing methods because it well handles the noise in our idea and in practice.
In other words, SDProp is expected to effectively train the model even if we use small mini-batch sizes that incur noisy first order gradients \cite{dekel2012optimal}.
Therefore, we investigated the sensitivity of SDProp and existing methods to mini-batch size.
While the main purpose of this experiment is to reveal the one performance attribute of SDProp, the result suggests that SDProp can be used on devices with scant memory that must use small mini-batches.

We compared SDProp to RMSProp and Adam using mini-batch sizes of 16, 32, 64 and 128.
We used the Cifar-10 dataset for the 10-class image classification task.
We used CNN as per the previous section.
The hyper-parameters are also the same as the previous section; they are tuned by grid search.
The number of epochs was 50.

Table \ref{batchsize} shows the final training accuracies.
SDProp outperforms RMSProp and Adam in all mini-batch size values examined. 
Specifically, although small mini-batch size of 16 incurs very noisy first order gradients, SDProp obviously achieves effective training unlike RMSProp and Adam.
In addition, Table \ref{batchsize} shows that the superiority of our approach over RMSProp and Adam increases as mini-batch size falls.
For example, if the mini-batch size is 16, our approach has 8.75 percent higher accuracy than RMSProp and 2.24 percent more accurate if the mini-batch size is 128.
This indicates that our covariance matrix based preconditioning effectively handles the noise of first order gradients.

\subsection{Efficiency and Effectiveness for RNN}
We evaluated the efficiency and effectiveness of SDProp for the Recurrent Neural Network (RNN).
In this experiment, we predicted the next character by using previous characters via character-level RNN.
We used the subset of shakespeare dataset and the source code of the linux kernel as the dataset \cite{karpathy2015visualizing}.
The size of the internal state was 128.
The pre-processing of the dataset followed that of \cite{karpathy2015visualizing}.
The mini-batch size was 128.
In SDProp, we tried grid search with \(\rho \in \{0.1, 0.01, 0.001\}\) and \(\gamma \in \{0.9, 0.99\}\).
As a result, SDProp used the settings of \(\rho=0.01\) and \(\gamma=0.99\).
In RMSProp, we tried grid search with \(\alpha \in \{0.1, 0.01, 0.001\}\) and \(\beta \in \{0.9, 0.99\}\).
Finally, we used the settings of \(\alpha=0.01\) and \(\beta=0.99\) for RMSProp.
The training criterion was cross entropy.
We used gradient clipping and learning rate decay.
Gradient clipping is a popular approach for scaling down the gradients by manually setting a threshold; it prevents gradients from exploding in RNN training \cite{DBLPPascanuMB13}.
We set the threshold to 5.0.
We decayed the learning rate \(\alpha\) every tenth epoch by the factor of 0.97 for RMSProp following \cite{karpathy2015visualizing}.
In SDProp, \(\rho\) was also decayed the same as \(\alpha\) of RMSProp.

\begin{figure}[t]
\begin{center}
 \begin{minipage}{0.41\hsize}
  \begin{center}
\includegraphics[viewport = 0.000000 0.000000 504.000000 504.000000, width=\columnwidth]{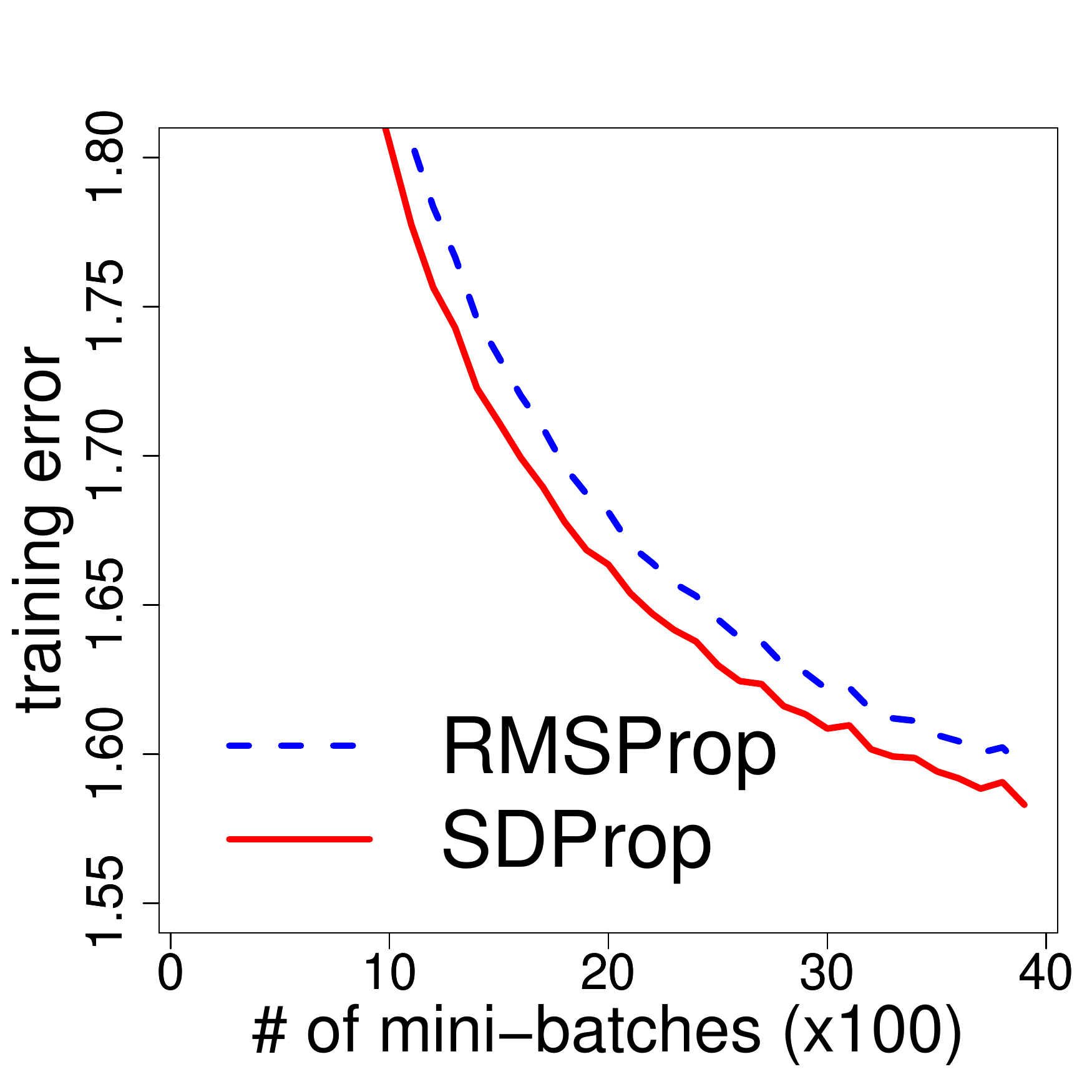} \\
  \end{center}
 \end{minipage}
 \begin{minipage}{0.41\hsize}
  \begin{center}
\includegraphics[viewport=0.000000 0.000000 504.000000 504.000000, width=\columnwidth]{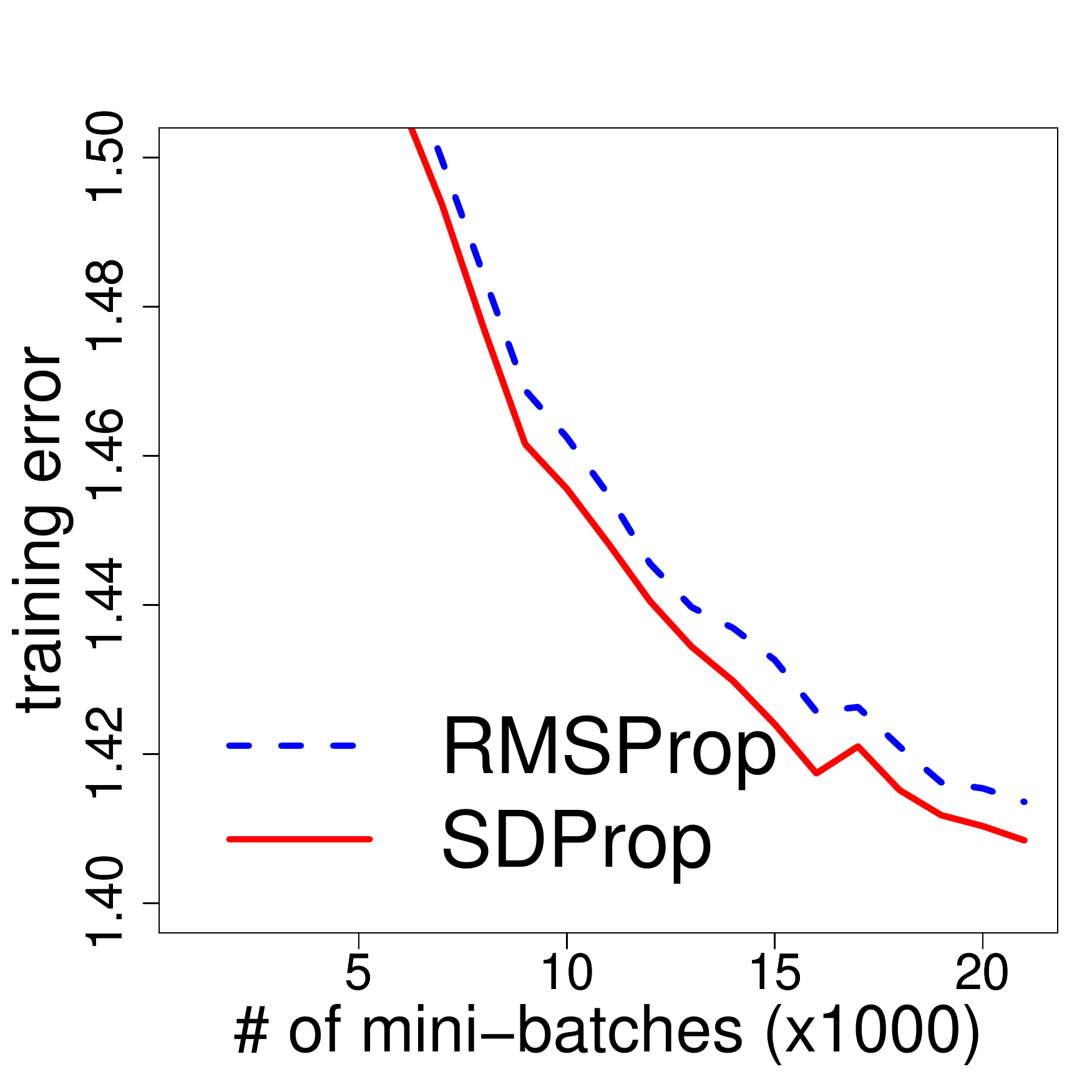} \\
  \end{center}
 \end{minipage}
 \caption{\small Cross entropies in training RNN for shakespeare dataset (left) and source code of linux kernel (right).}
 \label{SDProp_rnn}
\end{center}
\end{figure}

Figure \ref{SDProp_rnn} shows the results of the shakespeare dataset and the source code of the linux kernel.
SDProp reduces the training loss faster than RMSProp.
Since SDProp effectively handles the noise induced by the mini-batch setting, it can efficiently train models other than CNN, such as RNN.

\subsection{20 Layered Fully-connected Neural Network}
\begin{table}[!t]
  \caption{\small Average, Best and Worst training accuracy percentage of 20 layered fully-connected networks.} 
  \vskip 0.1in
  \label{simpleinit}
  \centering
  \begin{tabular}{|c|c|c|c|c|c|} \hline
    \multicolumn{3}{|c|}{} & \multicolumn{3}{|c|}{Accuracy} \\ \hline
    Method & \(\alpha\) & \(\beta\) & Ave. & Best & Worst \\ \hline 
     RMSProp & 0.001& 0.9 & 92.86 & 97.95 & 84.79 \\ \cline{3-6}
     & &0.99 & 98.81 & 99.11 & 98.34 \\ \hline \hline
     & \(\rho\) & \(\gamma\) & Ave. & Best & Worst \\ \hline 
     SDProp& 0.001& 0.9 & 93.77 & 97.9 & 87.57  \\ \cline{3-6}
     & & 0.99 &  \(\textbf{99.20}\) & \(\textbf{99.42}\) & \(\textbf{99.09}\) \\ \hline
  \end{tabular}
\end{table}

In this section, we performed experiments to evaluate the effectiveness of SDProp for training deep fully-connected neural networks.
\cite{NIPS2014_5486} suggests that the number of saddle points exponentially increases with the dimensions of the parameters.
Since deep fully-connected networks typically have parameters with higher dimension than other models such as CNN, this optimization problem has many saddle points.
This problem is challenging because SGD slowly progresses around saddle points \cite{NIPS2014_5486}.

We used a very deep fully-connected network with 20 hidden layers, 50 hidden units and ReLU activation functions.
We used the MNIST dataset for the 10-class image classification task.
This setting is the same as \cite{neelakantan2015adding} used in evaluating the effectiveness of SGD with high dimensional parameters.
Note that MNIST is sufficient for our evaluation because, unlike CNN, fully-connected networks do not saturate the accuracy in our experiment.
Our purpose is to evaluate the effectiveness under the setting of very high dimensional parameter.
Thus, it is sufficient to evaluate effectiveness if the accuracy is not saturated.
The training criterion was negative log likelihood.
The mini-batch size was 128.
We initialized parameters from a Gaussian with mean 0 and standard deviation 0.01 following \cite{neelakantan2015adding}.
We compared SDProp to RMSProp.
In SDProp, we tried the combinations of hyper-parameters by using \(\gamma \in \{0.9, 0.99\}\) and \(\rho \in \{0.1, 0.01, 0.001\}\).
In RMSProp, we tried the combinations of hyper-parameters by using \(\beta \in \{0.9, 0.99\}\) and \(\alpha \in \{0.1, 0.01, 0.001\}\).
The number of epochs was 50.
Although these algorithms are trapped around saddle points, its frequency may depend the initialization of parameter.
Therefore, we tried 10 runs for each of the above settings.

Table \ref{simpleinit} lists the results for the best setting of \(\alpha\) and \(\rho\).
It shows averages, best, worst of training accuracies for each setting.
The result shows that SDProp achieves higher accuracy than RMSProp for the best setting.
In addition, the difference between best and worst accuracy of SDProp is smaller than RMSProp.
Since SDProp effectively handles the randomness of noise, it can reduce result uncertainty.
The results show that SDProp effectively trains models that have very high dimensional parameters.

\section{Conclusion}
\label{sec:conclusion}
We proposed SDProp for the effective and efficient training of deep neural networks.
Our approach utilizes the idea of using covariance matrix based preconditioning to effectively handle the noise present in the first order gradients.
Our experiments showed that, for various datasets and models, SDProp is more efficient and effective than existing methods.
In addition, SDProp achieved high accuracy even if the first order gradients were noisy.

\bibliographystyle{unsrt}
\bibliography{main}

\end{document}